\documentclass[]{fairmeta}
\usepackage{makecell}
\usepackage{wrapfig}
\usepackage{tabularx}
\usepackage{textcomp}
\usepackage{stfloats}
\usepackage{url}
\usepackage{verbatim}
\usepackage{titlesec}
\usepackage{tocloft}
\usepackage{adjustbox}
\usepackage{multirow}
\usepackage{pifont}
\usepackage[sc]{mathpazo}
\usepackage{tikz}
\usepackage{comment}
\usepackage{amsmath,amssymb}
\usepackage{colortbl}
\usepackage{natbib}
\usepackage{color}
\usepackage{booktabs} 
\usepackage{hyperref}
\usepackage{graphicx}
\usepackage{subcaption}
\RequirePackage{xspace}
\makeatletter
\DeclareRobustCommand\onedot{\futurelet\@let@token\@onedot}
\def\@onedot{\ifx\@let@token.\else.\null\fi\xspace}
\usepackage[most]{tcolorbox}
\usepackage{array}
\usepackage{siunitx}
\usepackage[table]{xcolor}
\usepackage{caption}
\definecolor{headerpurple}{HTML}{d8d2fc}
\definecolor{rowgray}{gray}{0.95}

\makeatother

\definecolor{adptorange}{RGB}{248, 205, 172}
\definecolor{cmpblue}{RGB}{189, 215, 238}

\definecolor{our_red}{RGB}{232,157,160}
\definecolor{our_blue}{RGB}{136,206,230}
\definecolor{our_orange}{RGB}{246,200,168}
\definecolor{our_green}{RGB}{178,211,164}

\definecolor{attn_code0}{RGB}{247,215,200}
\definecolor{attn_code1}{RGB}{238,169,139}
\definecolor{mlp_code0}{RGB}{204,201,221}
\definecolor{mlp_code1}{RGB}{102,95,153}
\definecolor{mygray}{HTML}{f0f0f0}

\definecolor{token_blue}{RGB}{84, 120, 140}

\usepackage{bbding}
\usepackage{fontawesome}
\usepackage{float}

\newlength\savewidth

\newcolumntype{x}[1]{>{\centering\arraybackslash}p{#1pt}}
\newcolumntype{y}[1]{>{\raggedright\arraybackslash}p{#1pt}}
\newcolumntype{z}[1]{>{\raggedleft\arraybackslash}p{#1pt}}

\renewcommand{\paragraph}[1]{\vspace{1.25mm}\noindent\textbf{#1}}

\usepackage{algorithm}
\usepackage{listings}

\definecolor{codeblue}{rgb}{0.25, 0.5, 0.5}
\definecolor{codekw}{rgb}{0.35, 0.35, 0.75}
\lstdefinestyle{Pytorch}{
    language = Python,
    backgroundcolor = \color{white},
    basicstyle = \fontsize{9pt}{8pt}\selectfont\ttfamily\bfseries,
    columns = fullflexible,
    aboveskip=1pt,
    belowskip=1pt,
    breaklines = true,
    captionpos = b,
    commentstyle = \color{codeblue},
    keywordstyle = \color{codekw},
}

\definecolor{green}{HTML}{009000}
\definecolor{red}{HTML}{ea4335}

\title{AlayaWorld: Long-Horizon and Playable Video World Generation}
\author{AlayaWorld Team, Alaya Lab}

\abstract{
Game worlds have traditionally been built through labor-intensive production pipelines, making them costly to develop, difficult to customization, and expensive to modify after deployment. Recent advances in video world models offer a fundamentally different paradigm. Rather than explicitly authoring every component of a virtual environment, these models autoregressively synthesize future observations conditioned on the current world state and user interactions, enabling playable worlds to be generated online. Trained on both gameplay recordings and real-world videos, they can capture diverse visual appearances and physical dynamics, opening new opportunities for interactive applications beyond gaming, including embodied intelligence. In this paper, we present \textbf{AlayaWorld}, a full-stack open-source framework for building interactive generative worlds. AlayaWorld enables open-ended real-time interaction, allowing users to freely navigate and perform diverse actions such as combat, spell casting, and monster summoning. The framework unifies the complete development-from data preparation model architecture, model training, inference acceleration, and deployment-within a modular and extensible architecture. Alongside the framework, we release reproducible pipelines, reference implementations, evaluation tools, and comprehensive documentation, establishing a practical foundation for future research and real-time applications of generative world models.
}

\github{\url{https://alaya-lab.github.io/AlayaWorld/}}
\video{\url{https://www.youtube.com/watch?v=n0jIEg7taTI}}
\correspondence{kaipeng.zhang@shanda.com}
\date{\today}
\begin{document}

\maketitle
\begin{figure}[!h]
    \centering
    \vspace{-1.8cm}
    \includegraphics[width=\linewidth]{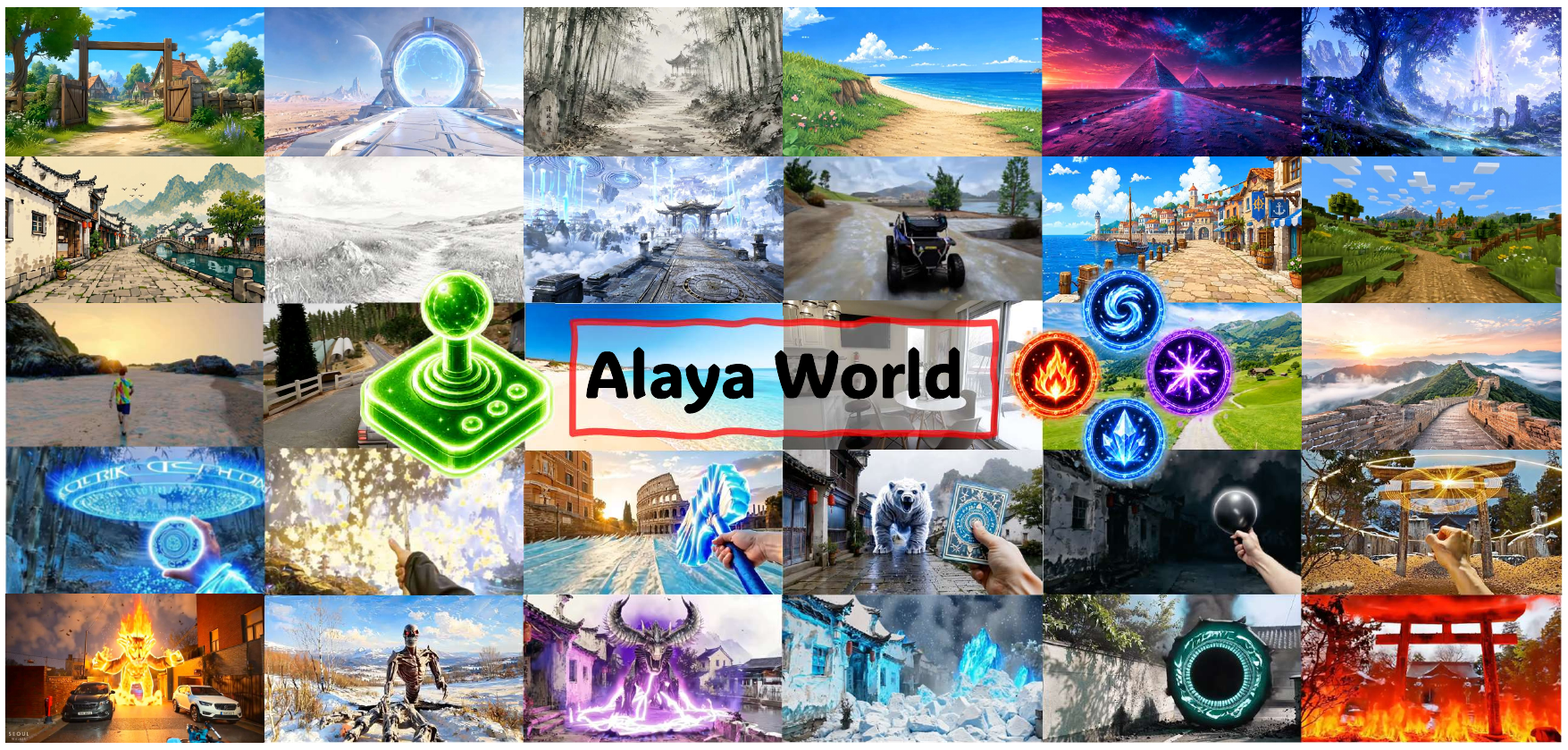}\vspace{-1.8cm}
    \captionof{figure}{\textbf{Interactive world simulation across diverse scenes.}
        AlayaWorld synthesizes explorable worlds that span first- and third-person viewpoints, real-world, game, and synthetic domains, and both indoor and outdoor environments. Moreover, it accommodates open-ended actions such as spell-casting, weapon combat, and monster summoning.}
    \label{fig:teaser}
\end{figure}

\section{Introduction}

\textbf{Alaya represents the foundation where intelligence, creativity, and new worlds are born.}

Interactive virtual worlds lie at the core of 3D video games and are increasingly serving as experimental platforms beyond entertainment, including embodied agent, robotics simulation, and the study of human decision-making in controllable environments.
The significance of these worlds stems not only from their visual realism but also from their interactivity: players or agents continuously issue actions, and the environment responds with coherent and persistent streams of observations.
Consequently, constructing virtual worlds that are simultaneously rich, interactive, and persistent has long been a central objective in artificial intelligence.

Conventional game world creation has relied on a labor-intensive production pipeline.
While it provides high-quality game, it comes at substantial cost. Objects, animations, gameplay, interaction rules and others must all be explicitly specified in advance, resulting in worlds that are largely predefined and difficult to modify after deployment. Extending existing environments or introducing new content often requires re-engaging the entire production process, limiting both the scalability and adaptability of virtual worlds.

Recent advances in Video world models~\cite{alibaba2026happyoyster, ball2025genie3, echo2026joyai, Mao_2026_CVPR, mao2025yume, team2026advancing, hunyuanvideo2025, wang2026matrix, yuan2026helios, zhu2026sana} have introduced a fundamentally different paradigm for interactive world creation.
Rather than explicitly constructing the whole world, generative models can directly predict future observations and synthesize subsequent visual states conditioned on user interactions.
Under this paradigm, a single model implicitly performs content generation, behavior modeling, and rendering, learning the underlying regularities of the world directly from data rather than relying on manually authored rules.
Moreover, because these models can be trained on diverse videos, the environments they generate are no longer constrained by gaming.
Instead, they can exhibit visual appearances and controllable dynamics and physics, moving from engineered worlds to generated worlds and benefiting embodied intelligence.

However, building a playable world with video world models still suffers from challenges.
The first is \textit{control}, which asks how much of the world is actually open to players. Whether navigation is endless and whether action is arbitrary, unbounded by preset physical laws.
The second is \textit{consistency}, which asks how much of the world achieves spatial and temporal consistency with natural dynamics that remain physically plausible. 
The third is \textit{stability}, which asks whether it can do long-horizon generation without visual drift.
The fourth is \textit{runtime}, which asks whether it can achieve real-time generation at low latency.

In this work, we first review the major challenges and representative approaches in interactive generative world modeling, providing a unified perspective on a rapidly emerging research area.
Building upon this analysis, we introduce \textbf{AlayaWorld}, an autoregressive DiT integrated with a prompt-switching mechanism, an AdaLN-style camera-control module, a 3D cache, a history-compression module, an error bank, and few-step distillation for the above challenges.

AlayaWorld is a full-stack, open-source, and long-term project intended to serve as a foundation for future work on video world models. The complete technical details, experimental results, and full codebase will be released in mid-July.
\section{Related Work}

We first review the video generation models that serve as backbones for world generation, and subsequently discuss interactive world models that formulate world generation as an end-to-end, playable system.
A fine-grained, technique-level comparison is deferred to the corresponding subsections of \S\ref{sec:alayaworld}.

\subsection{Video Genration Models}

Modern video generation is built upon diffusion and latent-diffusion models~\cite{rombach2022high}, in which transformer-based denoisers (DiT)~\cite{peebles2023scalable} are scaled to large generators.
Sora~\cite{brooks2024video} established the ``video as world simulator'' paradigm, and subsequent open-source backbones, including Open-Sora~\cite{zheng2024open}, Open-Sora-Plan~\cite{lin2024open}, HunyuanVideo~\cite{kong2024hunyuanvideo}, CogVideoX~\cite{yang2025cogvideox}, Wan~\cite{wan2025wan}, LTX-Video~\cite{hacohen2024ltx}, Step-Video-T2V~\cite{ma2025step}, and MAGI-1~\cite{teng2025magi}, have made large-scale text- and image-to-video synthesis broadly accessible, complemented by strong proprietary systems such as Veo~\cite{google2024veo}, Kling~\cite{kuaishou2024kling}, Gen-3~\cite{runway2024gen3}, and MovieGen~\cite{polyak2024movie}.
A parallel line of work develops video foundation models for physical and embodied settings, including Cosmos~\cite{agarwal2025cosmos} and long-context video-language modeling~\cite{liu2025world}.
These models synthesize high-fidelity, temporally coherent video along a \emph{predetermined} trajectory and thus provide a natural backbone for interactive world generation.
These models provide the generative backbones upon which interactive world generation is built; how they are extended into interactive, user-driven worlds is the focus of the systems discussed below and of AlayaWorld.

\subsection{Interactive World Models}

An expanding line of research formulates world generation as an end-to-end, playable system.
Action-conditioned world models such as Genie~\cite{bruce2024genie} learn controllable environments from large-scale unlabeled video, and Genie~2~\cite{parker2024genie} extends this formulation to explorable 3D scenes.
A related direction employs diffusion or autoregressive models as neural game engines that directly simulate playable worlds, including GameNGen~\cite{valevski2025diffusion} on DOOM, DIAMOND~\cite{alonso2024diffusion}, Oasis~\cite{decart2024oasis} on Minecraft, and promptable or playable game models~\cite{menapace2024promptable,yang2024playable}; GameGen-X~\cite{che2025gamegen} further addresses open-world game video generation with a dedicated dataset.
As many of these systems exhibit limited generalization beyond a single game, subsequent work seeks to improve controllability and generalization, exemplified by The Matrix~\cite{feng2026matrix}, GameFactory~\cite{yu2025gamefactory}, and the open-source real-time MineWorld~\cite{guo2025mineworld}.
More aligned with free-exploration interactive world generation, Yume~\cite{mao2025yume} synthesizes an explorable world from a single image under keyboard-style action control, and its successor Yume~1.5~\cite{Mao_2026_CVPR} further enhances generation quality and interactive controllability.
Recent systems increasingly target real-time, streaming interaction with extended consistency: Genie~3~\cite{ball2025genie3} generates navigable environments in real time with minute-level coherence but remains closed-source; Matrix-Game~\cite{he2025matrix} and Hunyuan-GameCraft~\cite{li2025hunyuan} integrate action conditioning, history-based consistency, and few-step distillation to achieve real-time generation with publicly released weights, and are further extended with long-horizon memory and instruction-based control~\cite{wang2026matrix,tang2025hunyuan}; and Lingbot-World~\cite{team2026advancing} improves long-horizon geometric consistency by scaling the context length of the diffusion model.
In a complementary direction, gaming-oriented efforts such as Microsoft's WHAM/Muse~\cite{kanervisto2025world} and its real-time variant WHAMM~\cite{microsoft2025whamm} jointly model environments and human actions for playable generation.
\section{AlayaWorld}
\label{sec:alayaworld}

AlayaWorld is a full-stack framework for interactive generative worlds. It is fine-tuned from LTX-2.3 combined with our designed modules.

\subsection{Interaction}
\label{subsec:agency}

AlayaWorld provide two types of interaction, including navigation and prompt-driven action.

\subsubsection{Navigation} 
\textbf{Camera as a conditioning signal.}
A first line of work treats camera motion as an external condition that the video model must translate into image-space motion.
Training-free methods steer pretrained video models at inference time by manipulating intermediate latents, attention features, or layout priors.
CamTrol~\cite{hou2024camtrol} models camera motion through 3D point-cloud rearrangement and uses the layout prior of noisy latents to guide generation without additional camera-pose training.
Latent-Reframe~\cite{zhou2025latentreframe} follows a related principle by reframing latent features through 3D point-cloud remapping.
These methods are compatible with off-the-shelf video generators, but their control accuracy is limited by the quality of the estimated geometry and by the indirect nature of latent steering.
Learned camera-conditioning methods instead train dedicated modules to encode camera trajectories.
MotionCtrl~\cite{wang2024motionctrl} represents camera motion as a sequence of rotations and translations and injects it through a camera motion control module in temporal transformer blocks.
CameraCtrl~\cite{he2025cameractrl} represents camera poses with Plücker ray embeddings and fuses the encoded camera features into temporal attention layers.
CamI2V~\cite{zheng2024cami2v} and CamCo~\cite{xu2024camco} further strengthen this line by combining Plücker-based pose conditioning with explicit geometric constraints such as epipolar attention.
CameraCtrl II~\cite{he2025cameractrl2} extends camera-conditioned video diffusion to dynamic scene exploration over broader viewpoint ranges.
Another variant discretizes camera motion into action tokens.
Yume~\cite{mao2025yume} exposes camera control through keyboard-style actions, which simplifies interactive use but provides only action-level rather than arbitrary continuous trajectory control.
Overall, these methods provide flexible camera control through continuous poses, discrete actions, or latent-space steering, but their accuracy depends on how reliably the generator learns to translate the condition into consistent image-space motion.

\textbf{Camera as an architectural bias.}
A second line of work incorporates camera geometry into the internal operators of the transformer.
Ray-map conditioning provides pixel-aligned camera rays as token-level geometric inputs.
Relative camera encodings instead make token interactions depend on the geometric relationship between viewpoints.
PRoPE~\cite{li2025prope} encodes complete camera frustums, including intrinsics and extrinsics, as a projective relative positional encoding inside attention.
HY-World~1.5~\cite{hyworld2025} adopts this idea for interactive world modeling by injecting continuous camera poses into causal self-attention through PRoPE, while incorporating discrete keys into the timestep embedding for robust user control.
UCPE~\cite{zhang2025ucpe} further generalizes camera positional encoding by modeling 6-DoF poses, camera intrinsics, lens distortion, and absolute orientation within a unified formulation.
Other works use normalization-level modulation to inject camera information with minimal overhead.
BulletTime~\cite{wang2025bullettime} combines camera and temporal control through a lightweight 4D positional encoding together with a camera-conditioned normalization branch (Camera-AdaLN), where the Camera-AdaLN branch modulates DiT features through camera-dependent affine normalization parameters.
These architectural mechanisms are efficient and compatible with modern transformer-based video generators.
However, because camera control is injected into the denoising backbone, weak conditioning may be ignored, while overly strong or miscalibrated conditioning can degrade appearance quality and temporal coherence.

\textbf{Camera as explicit rendered evidence.}
A third line of work reduces the burden of implicit pose-to-image reasoning by rendering a geometric proxy under the target camera trajectory.
GEN3C~\cite{ren2025gen3c} constructs a 3D cache from depth-unprojected seed frames or previously generated frames, renders this cache along the desired trajectory, and conditions the video generator on the rendered view.
This converts camera control from a parameter interpretation problem into an image-conditioned generation problem.
The generator can then focus on completing disoccluded regions, correcting rendering artifacts, and advancing scene dynamics.
TrajectoryCrafter~\cite{yu2025trajectorycrafter} follows a related generative re-rendering formulation by jointly conditioning on point-cloud renders and source videos to redirect camera trajectories for monocular videos.
ReCamMaster~\cite{bai2025recammaster} and ReCapture~\cite{zhang2024recapture} re-render a given video under a novel camera trajectory while preserving scene appearance and dynamics.
CamCloneMaster~\cite{luo2025camclonemaster} transfers camera motion from a reference video without requiring explicit camera parameters at inference time.
We note that this third line operates in a video-to-video (V2V) re-rendering setting: it takes an existing source video as input and redirects its camera trajectory, and is therefore distinct from the interactive free-exploration setting that synthesizes an unobserved world from a single frame or a text prompt.
These methods provide direct image-space guidance for the target viewpoint, which improves geometric fidelity and camera-following accuracy.
Their cost is a heavier pipeline that depends on geometry estimation, rendering quality, source-video coverage, or reference-motion availability.

\textbf{Our approach.}
AlayaWorld combines explicit rendered evidence with lightweight architectural injection.
Following GEN3C~\cite{ren2025gen3c}, we maintain a 3D cache and render it along the player's target camera trajectory.
The rendered cache gives the generator concrete visual evidence for the queried viewpoint, improving trajectory following and cross-view consistency.
This is particularly important for interactive worlds, where the player may leave a region and later return to it.
At the same time, we avoid heavy camera-fusion modules inside the generative backbone.
We inject the compact camera condition through AdaLN-style modulation, which introduces only a small parameter and computation overhead.
This design separates the roles of the two conditioning paths.
The rendered 3D cache provides spatially grounded appearance and geometry.
The lightweight camera modulation provides trajectory awareness inside the backbone.
Together, they support precise agency while preserving the efficiency required for responsive interaction.

\subsubsection{Prompt-driven Action} 
Navigation is a type of basic interaction while AlayaWorld also support freely prompt-driven actions, such as spell-casting, weapon combat, and monster summoning, to achive a real playable world. In particular, AlayaWorld introduce a prompt switching mechanism at chunk granularity. It can replace a text condition at any chunk boundary, so that newly generated content from the next chunk does not affect previously generated content, avoiding re-generation of the existing sequence. This mechanism is conceptually aligned with attention-level prompt editing~\cite{hertz2022prompt2prompt,liu2024videop2p}.

\subsection{Consistency}
\label{subsec:persistence}

Consistency requires spatial and temporal consistency with natural dynamics that remain physically plausible. For example, a player leaves a region and later returns to it.
We emphasize that this is a property of place identity: even if the generated stream is perfectly stable and free of drift, persistence still fails whenever a revisited region looks inconsistent with how it appeared before. It is tested by loop-closing ``leave-and-return'' trajectories and is independent of raw visual quality.
The difficulty is structural: autoregressive world models condition each new segment on a bounded context, while attention cost grows with the number of retained frames.
Keeping the entire rollout is therefore computationally prohibitive, whereas keeping only a recent window discards the evidence required for loop closure.
We organize prior work by the indexing principle that determines how past evidence is retrieved: temporally indexed memory recalls history by when it was observed, whereas spatially indexed memory recalls history by where it was observed, which better matches the loop-closure nature of revisits.

\textbf{Temporally indexed memory.}
Temporally indexed methods represent history as a sequence and preserve past evidence according to recency, narrative order, or recurrent propagation.
A first line directly retains raw temporal context, either by appending past frames to the current generation window or by keeping selected keyframes across shots.
Such approaches preserve high-fidelity observations when the relevant evidence remains in context, but their cost grows with history and their recall weakens once a location falls outside the retained window.
A second line compresses the temporal context before conditioning the generator.
FramePack~\cite{zhang2025framepack} packs long histories into a bounded context budget by progressively compressing past frame information.
Frame Preservation~\cite{zhang2025framepreservation} learns a lightweight history encoder that maps long video histories to short embeddings while preserving frame-level information at arbitrary temporal positions.
Other compression-based methods use mixture, packing, memory-flow, or hierarchical memory designs to reduce the cost of long-range conditioning~\cite{zhang2025framepack,zhang2025framepreservation,wu2026infiniteworld}.
A third line carries history implicitly through recurrent or state-space computation.
These methods propagate a compact hidden state across segments rather than explicitly storing all visual tokens~\cite{chen2025recurrentdiffusion,po2025longcontextssm,yu2025videossm}.
They offer favorable scaling, but the compressed state is indexed by rollout order rather than by physical location.
Consequently, temporal memory can maintain short-term dynamics and global continuity, but it has no direct guarantee that the exact evidence needed for a long-delayed revisit will remain accessible.

\textbf{Spatially indexed memory.}
Spatially indexed methods organize history by viewpoint, pose, or reconstructed scene geometry.
A lightweight form stores historical frames together with camera states and retrieves them by spatial overlap.
Context-as-Memory~\cite{yu2025contextasmemory} keeps history in frame format and selects relevant context according to field-of-view overlap.
WorldMem~\cite{xiao2025worldmem} stores memory frames with poses and timestamps, then reads the memory according to the queried state.
This type of memory is still frame-based, but its retrieval key is spatial rather than purely temporal.
A stronger form binds memory to an explicit geometric substrate.
GEN3C~\cite{ren2025gen3c} builds a 3D cache from depth-unprojected frames and renders this cache under the target camera trajectory to condition future generation.
Video World Models with Long-term Spatial Memory~\cite{wu2025longtermspatialmemory}, EvoWorld~\cite{wang2025evoworld}, and Spatia~\cite{zhao2025spatia} maintain explicit spatial memories that are updated over the rollout and reprojected or queried when a viewpoint is revisited.
Lyra~2.0~\cite{shen2026lyra2} maintains per-frame 3D geometry for information routing, retrieving historical frames that are visible from the target view and establishing dense correspondences for long-horizon generation.
System-level 3D world models such as HY-World~2.0~\cite{hyworld22026} further combine view generation, reconstruction, world expansion, and composition to produce persistent navigable 3D scenes.
Recent work reduces the cost of explicit RGB-space spatial memory by storing the cache directly in diffusion latent space~\cite{zhao2025spatia}.
Spatial indexing is therefore better aligned with loop closure and long detours, because the queried viewpoint can retrieve evidence from the corresponding region rather than from the most recent segment.
Its main limitations are pipeline complexity, dependence on depth or geometry estimation, and the difficulty of representing dynamic objects whose state is not fixed in a static spatial cache.

\textbf{Our approach.}
These two forms of memory are complementary.
Spatial memory anchors the geometry and appearance of previously visited regions, while temporal memory captures recent motion, transient changes, and global rollout context.
AlayaWorld combines both mechanisms.
Following GEN3C~\cite{ren2025gen3c}, we maintain an explicit 3D cache and reproject it into the queried viewpoint, providing spatially grounded evidence for previously observed regions and improving consistency under revisits.
Because such a cache mainly represents static structure, it cannot by itself encode all recent temporal dynamics.
We therefore additionally compress the recent frame history into a lightweight embedding following Frame Preservation~\cite{zhang2025framepreservation}.
The explicit cache supplies spatial persistence, while the compressed history supplies temporal persistence.
Together, they cover the complementary failure modes of purely spatial and purely temporal memory.

\subsection{Stability}
\label{subsec:durability}

We instantiate \textit{stability} as \textit{long-horizon video generation}.
The goal is not only to generate more frames, but to maintain visual quality, object identity, motion continuity, and controllability as the rollout grows.
Unlike persistence, this is a property of \emph{stability over the horizon} and is independent of whether any region is revisited: even if the player never turns back and no loop closure ever occurs, a purely forward rollout still degrades as errors accumulate.
This is difficult because autoregressive video generation repeatedly conditions on its own outputs.
Small artifacts, inconsistent motion, or distribution shifts can therefore accumulate over time and push later segments away from the data manifold.
Interventions against this drift can be placed at four stages of the autoregressive pipeline: the conditioning input, the training procedure, the sampling schedule, and the prediction target.

\textbf{Conditioning input: longer and better-organized history.}
One way to mitigate drift is to give the generator access to longer or better organized history.
Unlike the spatial recall used for loop closure in \S\ref{subsec:persistence}, the concern here is how the conditioning input stabilizes the forward rollout, evaluated by degradation over a monotonic long rollout rather than by revisit consistency.
Some methods simply enlarge the usable context: Context Forcing~\cite{chen2026context} trains a long-context student with a matching teacher to avoid the short-context supervision mismatch.
Others compress or reorganize that context under a fixed budget: Infinite-World~\cite{wu2026infiniteworld} distills historical latents into a pose-free fixed-budget memory, while Relax Forcing~\cite{zhao2026relaxforcing} argues that adding memory alone is insufficient and instead assigns context to functional roles such as sink, tail, and selected history.
These methods keep more useful history active, but they still rely on the model to decide how that history should correct the current generation, so their effectiveness is bounded by memory selection, compression loss, and the training distribution of long contexts.
We note that, although they share mechanisms with the memory compression in \S\ref{subsec:persistence}, their goal is to stabilize the forward rollout rather than to enable loop-consistent recall via spatial indexing; the two are complementary rather than redundant.

\textbf{Training procedure: rollout-aware training and error correction.}
Another approach treats drift as a train-test mismatch.
Teacher-forced video models are trained on clean histories, but autoregressive inference conditions on self-generated frames that already contain errors.
Self-Forcing~\cite{huang2025selfforcing} addresses this by training the model under its own autoregressive rollout, reducing exposure bias between training and inference.
Stable Video Infinity~\cite{li2025stablevideoinfinity} makes this idea more explicit through Error-Recycling Fine-Tuning.
It injects historical errors into clean inputs, estimates the resulting residual errors, stores them in an error bank across diffusion timesteps, and resamples them during training.
This teaches the DiT to recognize and correct the types of errors it will encounter during long rollouts.
Helios~\cite{yuan2026helios} also targets long-video drift, but avoids runtime anti-drifting heuristics such as self-forcing, error banks, and keyframe sampling.
Instead, it characterizes drift failure modes and trains the model with simulated drifting histories while suppressing repetitive motion at its source.
These approaches directly address error accumulation, but they usually require additional long-rollout training, error simulation, or model-specific anti-drift recipes.

\textbf{Sampling schedule: relaxed-causal and anchor-based generation.}
Drift can also be addressed by changing the generation schedule so that errors do not propagate strictly from one segment to the next.
Rolling Forcing~\cite{liu2025rollingforcing} relaxes strict causality while still streaming forward, jointly denoising a rolling window of frames at increasing noise levels and using initial frames as attention sinks, which suppresses error growth at real-time latency.
Anchored Tree Sampling~\cite{bendel2026anchoredtreesampling} instead abandons causal order, generating sparse anchors over the full horizon and filling in intermediate spans, which bounds drift locally but requires non-causal generation and is therefore less suitable for real-time interaction.

\textbf{Prediction target: geometric and perceptual stabilizers.}
Drift can further be constrained by augmenting the prediction target with auxiliary signals that are more stable than raw RGB alone.
WorldWeaver~\cite{liu2025worldweaver} jointly models RGB frames and perceptual conditions, and uses depth cues as a memory signal because depth is less prone to visual drift than RGB appearance.
Endless World~\cite{zhang2025endlessworld} adds 3D-aware attention so that generation is constrained by explicit geometry and remains coherent over extended rollouts.
World models with explicit spatial memory also stabilize long rollouts by grounding generated views in persistent scene structure rather than relying only on previous RGB frames~\cite{wu2025longtermspatialmemory,wang2025evoworld,zhao2025spatia}.
These methods reduce drift by constraining generation with geometry, perception, or world-state cues.
Their cost is additional estimation, representation, or memory maintenance, and they may still struggle with dynamic objects whose state changes over long horizons.

\textbf{Our approach.}
AlayaWorld treats stability as a training-time robustness problem for autoregressive generation.
Following Helios~\cite{yuan2026helios}, we expose the model to drifted histories during training rather than assuming that every conditioning segment remains clean.
We further introduce an error bank that stores residual artifacts accumulated during rollout and reuses them as structured perturbations.
Unlike methods that apply such errors only to the predicted target, we inject error-bank samples into both the memory condition and the target segment.
This joint perturbation better matches long-horizon inference, where the model must generate from imperfect memory while also correcting errors in the next segment.
As a result, the model learns not only to continue a clean video, but also to stabilize generation under corrupted history and prevent errors from compounding across autoregressive steps.

\subsection{Runtime}
\label{subsec:responsiveness}

We instantiate \textit{runtime} as \textit{real-time video generation under interactive constraints}.
Interactive generation is bounded by two distinct latencies, and a responsive system must keep both small.
B\emph{Visual latency} is the time from deciding to generate to a frame appearing; it is a compute problem.
\emph{Semantic latency} is the time from a change in user intent to the output reflecting it; it is a conditioning-update problem.
We organize prior work accordingly: methods that make generation \emph{fast} reduce visual latency, while methods that make conditioning \emph{updatable on the fly} reduce semantic latency.

\textbf{Reducing visual latency: faster generation.}
Three complementary lines shorten the time from trigger to frame.
The first reduces the number of denoising steps through distillation.
Progressive Distillation~\cite{salimans2022progressive} and Consistency Models~\cite{song2023consistency} train a student to approximate the teacher's sampling trajectory or probability-flow solution, an idea moved into latent diffusion by Latent Consistency Models~\cite{luo2023lcm} and reformulated as distribution matching by DMD~\cite{yin2024dmd} and DMD2~\cite{yin2024dmd2}; VideoLCM~\cite{wang2023videolcm}, AnimateLCM~\cite{wang2024animatelcm}, and T2V-Turbo~\cite{li2024t2vturbo} extend few-step distillation to video.
The second restructures generation into a causal stream rather than an offline clip, so results are exposed continuously: Reuse and Diffuse~\cite{gu2023reusediffuse} and FIFO-Diffusion~\cite{kim2024fifodiffusion} continue generation by reusing latents or diagonal denoising, StreamingT2V~\cite{henschel2024streamingt2v} and StreamDiT~\cite{kodaira2025streamdit} stream over overlapping chunks or a moving buffer, and CausVid~\cite{yin2025causvid}, MotionStream~\cite{shin2025motionstream}, and LongLive~\cite{yang2025longlive} distill bidirectional teachers into causal students for frame- or chunk-level generation on the fly.
The third reuses computation at runtime: Pyramid Attention Broadcast~\cite{zhao2024pab} broadcasts redundant attention outputs across diffusion steps, and FasterCache~\cite{lv2024fastercache} dynamically reuses features and exploits redundancy between the conditional and unconditional branches of classifier-free guidance.
Distillation can weaken motion fidelity and diversity under aggressive step reduction; streaming introduces chunk-boundary discontinuity and error accumulation; and cache reuse only trims computation inside a fixed sampling process, leaving semantic runtime untouched.

\textbf{Reducing semantic latency: on-the-fly conditioning.}
Runtime also requires the conditioning state to change during rollout without regenerating the sequence.
The mechanism originates in attention-level prompt editing: Prompt-to-Prompt~\cite{hertz2022prompt2prompt} edits text-conditioned diffusion by manipulating cross-attention maps, and Video-P2P~\cite{liu2024videop2p} extends this to video while preserving temporal coherence in unchanged regions.
LongLive~\cite{yang2025longlive} makes this work under streaming by introducing KV-recache, which recomputes cached states with the new prompt and previously generated frames, removing semantic inertia while keeping visual continuity across a switch.
These methods show that runtime is not only sampling speed: the system must also update its conditioning quickly enough to follow user intent without abrupt visual changes or delayed semantic response.

\textbf{Our approach.}
AlayaWorld targets runtime through a simple real-time generation design.
We adopt standard DMD-based distillation to reduce the number of denoising steps required for each generated chunk.
We further use a small temporal chunk size so that each generation call has low latency and user commands can affect the output after a short delay.
Prompt switching is supported at chunk boundaries by updating the text condition before generating the next chunk.
This avoids full-sequence regeneration while keeping the interaction loop simple and predictable.
Compared with heavy runtime cache engineering or specialized KV-recache mechanisms, our design prioritizes low per-chunk compute and frequent condition update points.
In this way, few-step distillation provides the speed, short chunks provide the interaction granularity, and prompt switching provides controllability under changing user intent.
\section{Qualitative Results}

AlayaWorld is fine-tuned from LTX-2.3~\cite{hacohen2024ltx}, performing autoregressive generation at 720p 24fps, where each chunk is produced with four denoising steps and corresponds to roughly one second of video.
For fairness, all baselines are evaluated under the same input conditioning and resolution as AlayaWorld whenever their public implementations permit.

\subsection{Camera Control}
\begin{figure}[!h]
    \centering
    \includegraphics[width=\linewidth]{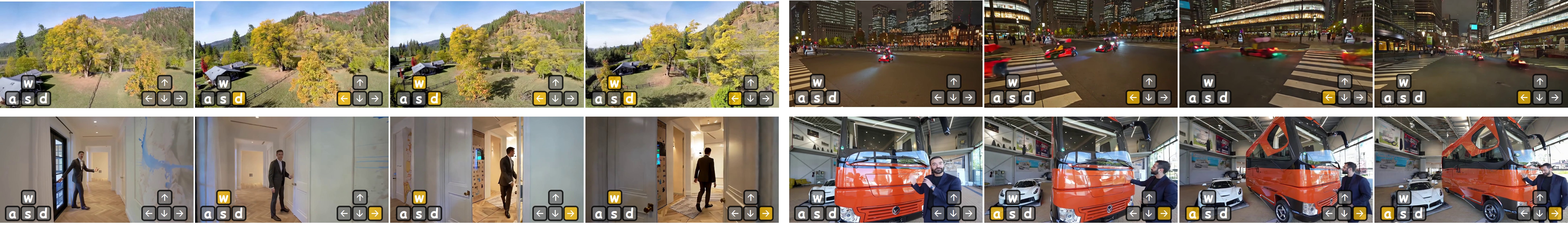}
    \captionof{figure}{\textbf{Qualitative results of camera-control generation.}}
    \label{fig:control}
\end{figure}

Figure~\ref{fig:control} shows generated sequences under varied camera and action commands.
AlayaWorld faithfully follows the requested viewpoint changes and translations while preserving scene identity and geometric plausibility, illustrating precise camera-controllable generation under interactive control.

\subsection{Open-ended Action}
\begin{figure}[!h]
    \centering
    \includegraphics[width=\linewidth]{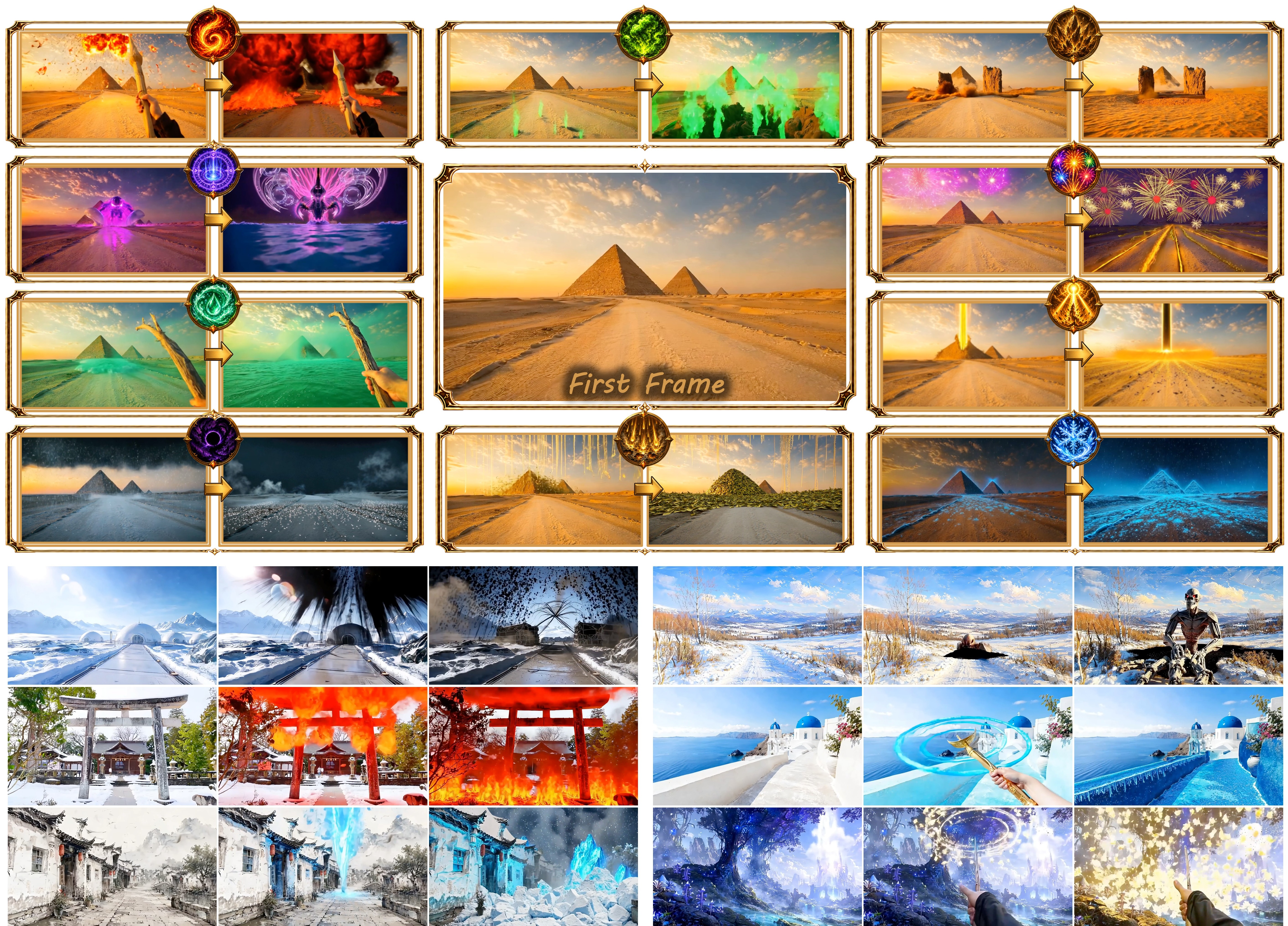}
    \captionof{figure}{\textbf{Qualitative results of prompt-driven actions.}}
    \label{fig:action}
\end{figure}

Figure~\ref{fig:action} shows sequences in which the text prompt is switched on the fly during generation.
By updating the text condition at chunk boundaries, AlayaWorld transitions to the new prompt within a short delay while preserving visual continuity across the switch, without regenerating the preceding sequence.
This illustrates that AlayaWorld follows changing user intent at interactive latency, supporting responsive control beyond a single fixed prompt to achieve open-ended actions.

\subsection{Consistency}
\begin{figure}[!h]
    \centering
    \includegraphics[width=\linewidth]{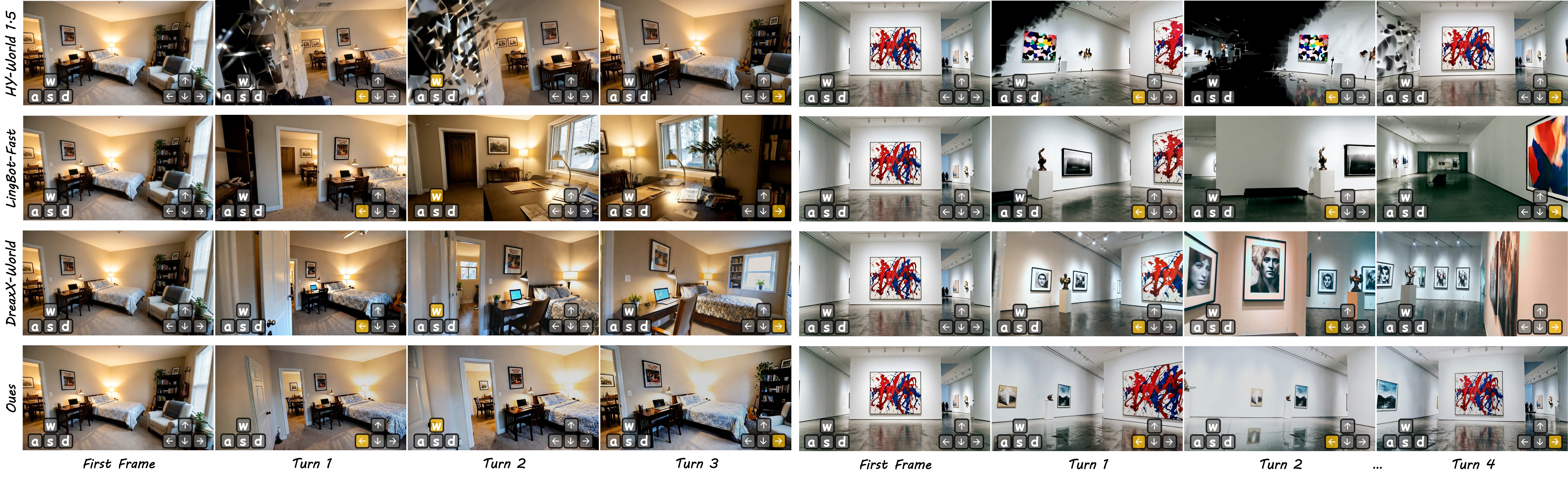}
    \captionof{figure}{\textbf{Consistency comparisons with existing method.}}
    \label{fig:memory}
\end{figure}

Figure~\ref{fig:memory} presents leave-and-return trajectories in which the viewpoint departs from a region and later revisits it.
The revisited regions remain consistent with their earlier appearance in geometry, layout, and texture, demonstrating that the explicit spatial cache and the compressed temporal history jointly support reliable loop closure.

We also evaluate representative interactive world models under the same setting. Prior models exhibit characteristic failure modes: visual degradation, inaccurate camera control, and inconsistency when previously observed regions are revisited. In contrast, AlayaWorld produces visually plausible and temporally coherent results that remain faithful to the control inputs while preserving scene structure across the trajectory.

\subsection{Long-Horizon Generation}
\begin{figure}[!h]
    \centering
    \includegraphics[width=\linewidth]{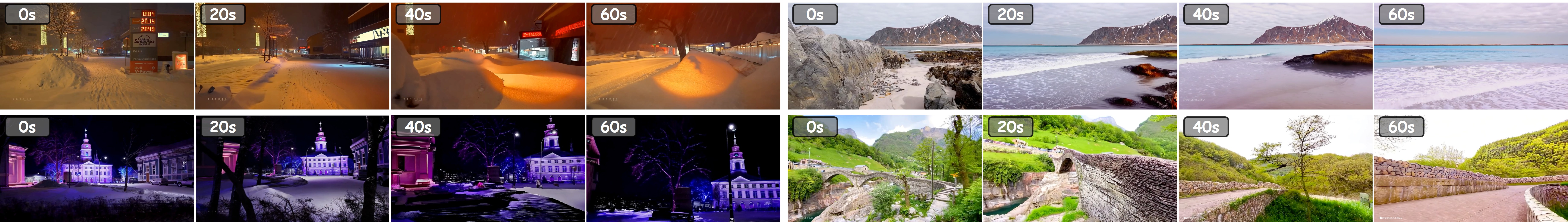}
    \captionof{figure}{\textbf{Qualitative results of one-minute generation.}}
    \label{fig:long}
\end{figure}

Figure~\ref{fig:long} shows extended rollouts generated autoregressively over long horizons.
AlayaWorld maintains visual quality, object identity, and motion continuity as the rollout grows, without pronounced accumulation of artifacts, indicating strong stability under purely forward exploration.

\subsection{Diverse Styles}
\begin{figure}[!h]
    \centering
    \includegraphics[width=\linewidth]{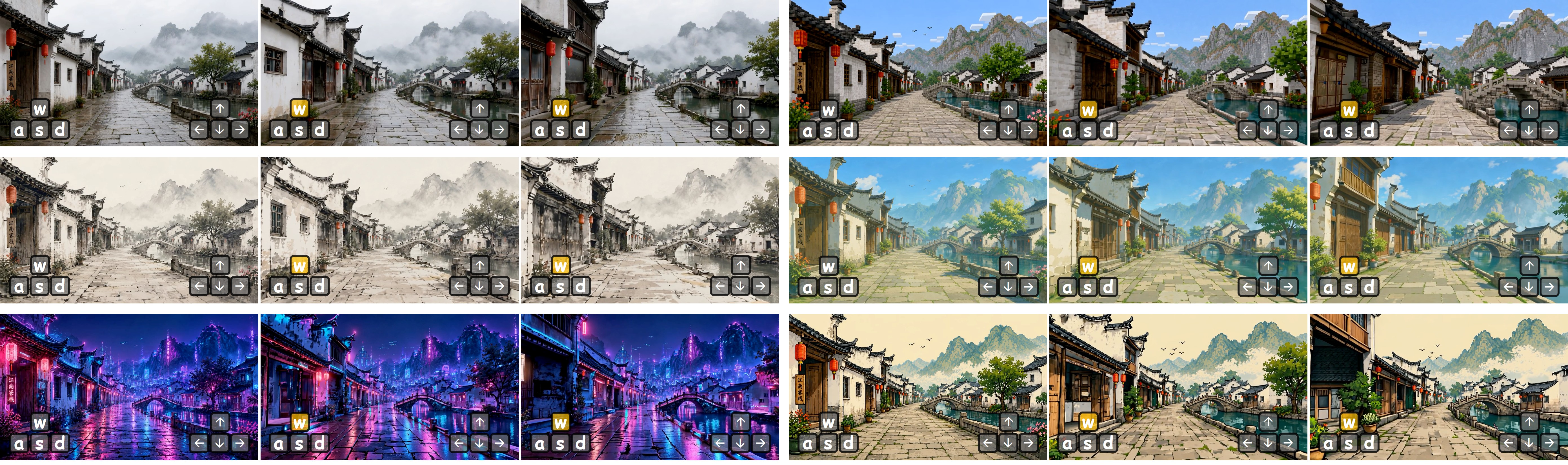}
    \captionof{figure}{\textbf{Qualitative results on diverse styles.}}
    \label{fig:style}
\end{figure}

AlayaWorld maintains scene consistency across a wide range of visual styles. As shown in Figure~\ref{fig:style}, the same navigation trajectory is rendered in realistic, Minecraft, ink painting, oil painting, cyberpunk, pixel art, and Zelda-inspired styles. While the rendering style varies significantly, the scene geometry, camera trajectory, and semantic content are consistently preserved.

\section{Contributions}

\textbf{Authors are listed in alphabetical order by their first names.}

\textbf{Core Lead:}
Kaipeng Zhang

\textbf{Lead:}
Chuanhao Li

\textbf{Core Contributor:}
Chuanhao Li, Kaipeng Zhang, Yifan Zhan, Yongtao Ge, Yuanyang Yin

\textbf{Contributor:}
Jiaming Tan, Kang He, Liaoyuan Fan, Ruicong Liu, Xiaojie Xu, Xuangeng Chu, Zhen Li, Zhengyuan Lin, Zhixiang Wang, Zian Meng, Zihui Gao

\bibliographystyle{abbrv}
\bibliography{references}

\end{document}